# Eyenet: Attention based Convolutional Encoder-Decoder Network for Eye Region Segmentation

Priya Kansal
Couger Inc.
Tokyo, Japan
priya@couger.co.jp

Sabari Nathan
Couger Inc.
Tokyo, Japan
sabari@couger.co.jp

## Abstract

*With the immersive development in the field of augmented and virtual reality, accurate and speedy eye-tracking is required. Facebook Research has organized a challenge, named OpenEDS Semantic Segmentation challenge for per-pixel segmentation of the key eye regions: the sclera, the iris, the pupil, and everything else (background). There are two constraints set for the participants viz MIOU and the computational complexity of the model. More recently, researchers have achieved quite a good result using the convolutional neural networks (CNN) in segmenting eye-regions. However, the environmental challenges involved in this task such as low resolution, blur, unusual glint and, illumination, off-angles, off-axis, use of glasses and different color of iris region hinder the accuracy of segmentation. To address the challenges in eye segmentation, the present work proposes a robust and computationally efficient attention-based convolutional encoder-decoder network for segmenting all the eye regions. Our model, named EyeNet, includes modified residual units as the backbone, two types of attention blocks and multi-scale supervision for segmenting the aforesaid four eye regions. Our proposed model achieved a total score of 0.974(EDS Evaluation metric) on test data, which demonstrates superior results compared to the baseline methods.*

## 1. Introduction

Recently, many disciplines are trying to integrate eye-tracking to automate their processes. For instance, to understand the attentional process [1][9] in patients, to study driver fatigue and drowsiness detection [7][8], to study the dynamics of consumers' attention for the deployment of an effective advertisement to promote a product [5][6] and for personal authentication over the smart devices using biometric security [2][3][4]. Other than that, one of the most significant users of the eye-tracking techniques is the augmented (AR) and virtual reality (VR) world. Most of the head-mounted video-based devices are gaze driven or eye-movement driven [10].

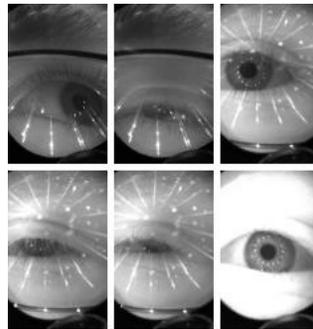

*Figure 1: Sample images from test data*

Also, the gaze-driven interaction schemes can enable novel methods of navigating and interacting with virtual environments [11] as well as in human-machine interaction [12].

To use eye-tracking in AR/VR devices, Facebook Research has organized on OpenEDS semantic Eye Segmentation Challenge [12], in which participants are asked to create a generalized model which is good to segment the four key eye regions within the constraint of least complexity in terms of model parameters.

To deal with the challenge presented in competition, we have proposed an Attention-based Convolutional Encoder-Decoder Network (EyeNet) to segment the boundaries of four different eye-regions (background, sclera, iris, pupil) and to sharply acquire the class pixels to ensure accurate segmentation of each region. EyeNet is based on exploiting residual connections for a better flow of information gradient using non-identity mapping (Non-IM) in both encoder and decoder. Furthermore, in pixel-wise segmentation tasks, it is important to emphasize the minute details of the boundary of the segmented region. Hence in EyeNet, we proposed two types of attention unit in the network. Also, we have also used multi-scale supervision to improve the segmentation accuracy. EyeNet is novel in the following four ways:

- It is an end-to-end semantic segmentation network for segmenting all the key eye-regions.

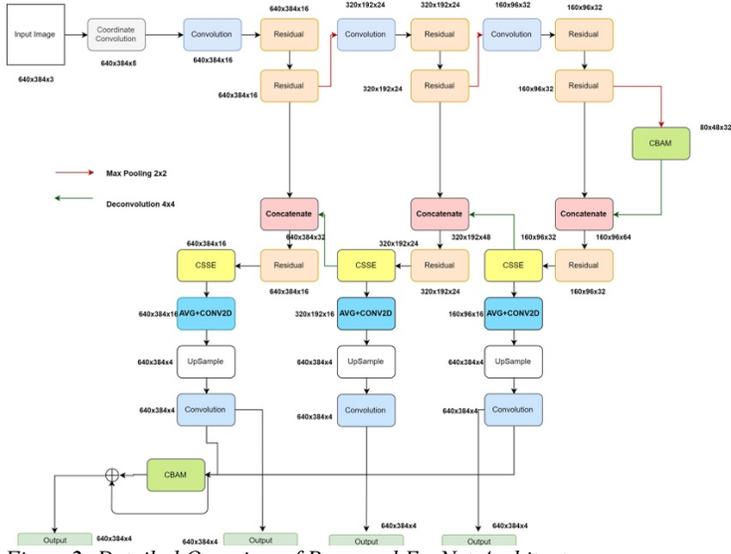
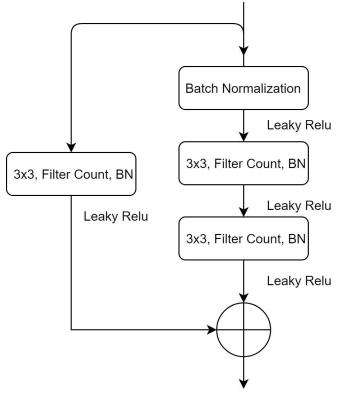

*Figure2: Detailed Overview of Proposed EyeNet Architecture*          *Figure3: Modified Residual Unit*

- It uses residual connectivities in both encoder and decoder to focus on the lost information without increasing complexity.
- Two attention units are used to get the accurate and sharp boundary of the segmented regions.
- EyeNet is computationally very less expensive when compared to other models.

The proposed EyeNet achieves new state-of-the-art performance on the data provided in the challenge. In particular, we achieved mean intersection over-union (mIOU) of 0.9494.

## 2. Related Work

Due to high paced development in AR/VR industry, eye-tracking and thus eye-region segmentation has become one of the most challenging research areas. Both the handcrafted and deep learning features are used to segment the eye-regions. However, handcrafted features are expensive, time taking also and requires lot of man hours. Since the emergence, the CNN based techniques have outperformed the traditional methods in every domain. Eye-region Segmentation is also not the exception for this. However, there are very few CNN based approaches which have focused on multi-class segmentation. [34] have proposed a CNN based multiclass model to segment five different classes of eye-region based on the encoder-decoder neural network model Seg-Net [35]. [36][37] have proposed the generative adversarial based approaches for sclera segmentation. [31] have proposed a residual unit based encode-decoder network for sclera segmentation and thus similar to our work as a backbone structure. Similarly, researchers are continuously working on FCNNs methods for iris segmentation e.g., [28][29][30][32][33]. These methods try to assign a predefined target label to each input pixel. [2] have used multi scale segmentation to segment the iris region. This multi scale supervision has inspired the current work also. Moreover, for pupil segmentation, [22][24][25][26][27] proposed some fully connected CNN based architecture and directly map the input images to the gaze results. In [23], researchers have first classified the eye-states, such as open, closed, half-opened etc. before applying specialized CNNs to estimate the pupil center coordinates, based on the eye state.

## 3. Method

### 3.1 Architecture Details

To deal with the challenge of robust multi class segmentation, we have proposed an attention-based convolutional encoder-decoder network. The detailed architecture is presented in Figure 2. In the following sub-section of this section, we will discuss the details of each unit we used in the proposed architecture.

**3.1.1 Modified Residual Unit**: Residual units have been a proven state-of-art approach for improving the accuracy in many domains [13][14]. In our modified residual block, the lost information of previous layer is again infused to the network using the non-identity mapping. This restored information can be used to enhance segmentation performance. We have used Leaky relu (0.1) as an activation function in the residual unit as inspired by [15]. Figure 3 shows the details of the Modified Residual unit.

**3.1.2 Convolutional Block Attention Module (CBAM):**
The CBAM block [16] is used to create a spatial attention on the channel attention of encoder output and the decoder output. Eq. (1) and Eq. (2) shows the details of mathematical operations.

Eq. (1)
$$C_A = f_a \left[ w_1 \left( w_0 \left( \frac{\sum_{i=1}^n x_i}{n} \right) \right) + w_1(w_0(\max(x_i))) \right]$$

where $w_0 \in R^{C/r \times C}$ and $w_1 \in R^{C \times C/r}$, $f_a$ is denotes the Sigmoid Activation

Eq. (2) $$S_A = f_{con}\left( \left[ \frac{\sum_{i=1}^n C_{Ai}}{n} || \max(C_{Ai}) \right] \right)$$

where, $f_{con}$ denotes the convolution operation

**3.1.3 Channel Squeeze and Spatial Excitation:** As inspired by [18] the output of the modified residual unit is further passed to the channel squeeze and spatial excitation (CS-SE) block [17] in each decoder block. The CS-SE block slices its input corresponding to the spatial location (x,y) where, x ∈ {1,2, ….H} and y ∈ {1,2, ….W}. This spatial mapping has helped the network in concentrating the meaningful features over the weak features.

**3.1.4 Multi-scale Supervision:** As the receptive field increases across successive layers, predictions computed at different layers embed spatial information at different scales. Especially for imbalanced multi-class segmentation, different scales can contain complementary information. In this paper, to increase the receptive field and avoid redundancy between successive scale predictions, average pooling and convolutions have been used for multi-scale supervision. As the gradient is propagated in intermediate layers, the network is able to update the weights more efficiently. For the purpose of final multi-class segmentation, these intermediate layers are fused.

As shown in Figure 2, first of all, the spatial coordinates of input are mapped with the coordinates in Cartesian space through the use of two extra channels (*i,j*) [19]. The coordinate channel *i* is a matrix in which row one is filled with all zeros, row two is all 1s, row three is all 2s and so on. Channel *j* is also similar to channel *i*, but in this, the columns are filled with zeros, 1s, 2s and so on. The output of coordinate convolutional layer with five channels is then fed to the encoder for down sampling. There are three down sampling blocks. In each block, we have used two modified residual units along with the convolutional layer. The down-sampled features are then fed to the CBAM block for attention. The concatenation of down-sampled features and the attention is fed into the decoder for up sampling. The decoder part consists of three up sampling blocks. The output of each up-sample block is fed to the residual block, CSSE block and then concatenated with corresponding encoder block output. Moreover, we have introduced the three side layers from each down sampling block, as suggested by [18] [20] [21]. The output of the CSSE block is passed to side layer, up sampled, average pooled, convolved and then supervised by the ground truth. The output of these three-side layers is then fused and supervised to get the segmentation mask. Dilation convolution is used for each convolution block in encoder (dilation ratio- 1,2,4) and decoder (dilation ratio- 4,2,1).

## 3.2 Dataset

The dataset provided in the OpenEDS Eye segmentation challenge consists of the eye images which are captured using a VR head-mounted device [12]. Out of 12759 images given in the training phase, 8916 images are used for training and rest 2403 images are used for validation. This is the same train and validation split, as suggested in the base paper [12]. Pixel wise annotation for eye-lid/sclera, iris, the pupil is provided. Some typical examples of these images are displayed in Fig.1.

## 3.3 Image Processing

Images are divided by 255 to normalize the value of each pixel between 0 and 1. Further, we have augmented some difficult to classify images, such as images with glasses, half-opened, dim/bright light using clipped zoom in (with 1.5 factor), Gaussian noise (mean = variance = 10) and rotation (from -10 degrees to 10 degrees). Total 8916 train images are augmented to 12000.

*Postprocessing* We have calculated the areas associated with the 8 nearest points by converting the segmented image to binary. The two largest areas corresponding to the background and eye region are selected for the next step. After careful analysis of segmented images, we found that the iris region has more noise. So, as the second step, the first step is repeated only for the iris region. Then we have combined all the channels and again have filled the holes of the iris region and the pupil region to get the final image.

## 3.4 Training

We have trained the network for four outputs which include the three side layers and one fused output layer for the mask of the input images. Prior to training, we have resized the images into 640x384. Adam optimizer is used to update the weights while training. The learning rate is initialized with 0.001 and reduced after 5 epochs to 10% if validation loss does not improve. The batch size is set to 2. The total epochs are set to 500. However, training is stopped early when the network started overfitting. The dataset is trained using Nvidia 1080 GTX GPU.

### 3.5 Evaluation Metrics

Model is evaluated using mean intersection-over union score (MIOU). However, other evaluation metrics such as Global and channel-wise pixel are also calculated to compare the results with the base model.

### 3.6 Loss Function

We have proposed loss function which is a sum of categorical cross-entropy loss and channel-wise dice loss [25] as defined in Eq. 3 and 4. The network is trained to minimize the SoftMax Dice loss.

$$DL = 1 - \frac{2\sum_{i=0}^{k} y_i p_i + \epsilon}{\sum_{i=0}^{k} y_i + \sum_{i=0}^{k} p_i + \epsilon} \quad Eq.(3)$$

$$L = -\sum_{j=0}^{M}\sum_{i=0}^{N} y_{ij}\log(p_{ij}) \quad Eq.(4)$$

where $y_{ij}, p_{ij}$ and are the ground truth and the predicted mask images respectively. The coefficient ε is used to ensure the loss function stability.

## 4. Results and Discussion

In this section, the detailed results and the comparison with state-of-art are given. Although the evaluation metric for the competition is MIOU, results are reported in terms of Pixel wise accuracy (PA), mean accuracy (MA), Dice value along with MIOU. Moreover, in this competition, the challenge was to create the least complex model. Hence, the no. of parameters in the model is also a reported to evaluate the model. Table1 presents the result comparison of EyeNet model and the baseline results in terms of all evaluation metrics.

Table 1: Comparison of Results of Base Paper and EyeNet

| Model | PA | MA | Dic | MIOU | M- Size | Para |
|---|---|---|---|---|---|---|
| mSegnet | 98.00 | 96.8 | - | 90.7 | 13.3 | 3.5 |
| mSegnet* | 98.3 | 97.5 | - | 91.4 | 13.3 | 3.5 |
| mSegnet** | 97.6 | 96.6 | - | 89.5 | 1.6 | 0.4 |
| ours (val) | 99.3 | 97.6 | 0.97 | 95.5 | 1.48 | 0.25 |
| ours (test) | - | - | - | 94.9 | 1.48 | 0.25 |

*\* mSegnet w/ BR; \*\* mSegnet w/ SC*

Further, the channel wise results on validation data are also presented in Table 2. The results show that our proposed model is quite accurate in the segmentation of each class. Fig. 4 shows the resulted segmentation mask of test data.

Table2: Channel wise Results on Validation set

| Class | Dice | Jaccard | IOU |
|---|---|---|---|
| Sclera | 0.96 | 0.93 | 0.96 |
| Iris | 0.98 | 0.96 | 0.98 |
| Pupil | 0.97 | 0.95 | 0.97 |

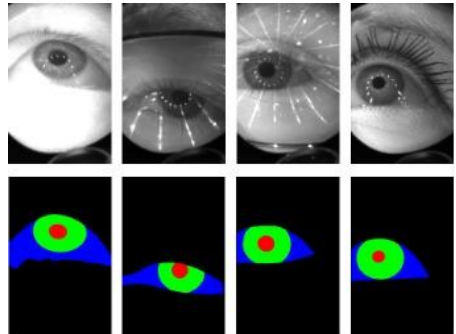

*Figure 4: Resulted Images of EyeNet Model*

Many experimentations have been done to improve the MIOU score in terms of changed backbone units of encoder and decoder, preprocessing, post-processing, using attention, image augmentation. Table 3 shows the results for some of these experimentations on validation data. Clearly, the proposed approach has achieved the highest score.

Table3: Experimentation Results

| Experiments | MIOU |
|---|---|
| Ours | **0.970** |
| without CBAM | 0.961 |
| without CSSE | 0.945 |
| without Coord layer | 0.961 |
| without Post-processing | 0.969 |

## 5. Conclusion and Proposed Future Work

In the present task, we tried to create a very less computationally complex model using the attention layer in the form of Eyenet. In Eyenet, we have leveraged a lot of proven techniques of accuracy improvement in a modified form (Coordinate convolutional layer, Residual units, Multi-Scale Supervision, loss function, etc.). As a future task, we would like to explore the other eye-regions like peri-ocular region, eye-lashes and medial canthus using this model as well as would like to train our model directly on low-resolution RGB images to improve the human-machine interaction in terms of improving machine intelligence about human emotion.

**Acknowledgement**: This work is supported by Couger Inc., Shibuya, Japan.